# A Picture Says Thousands of Words

## Harnessing Dermal Exposure Data from Images through Hybrid Deep Learning for Enhanced Safety Assessment

HUA QIAN, MANISHA KOTHA, TUAN TRAN, JENNIFER SHIN, AND HAINING ZHENG

Chemical handling tasks and product use patterns provide important contextual and behavioral information on exposure needed for assessment of workplace tasks and product safety. However, occupational health and safety professionals often are unable to observe all tasks performed by workers to document this type of information, nor is it practical to do so. It is even more difficult to obtain this data on consumers for product safety assessments. This difficulty leads OEHS professionals to rely on conservative assumptions or use historical data that may no longer be relevant; with continuous innovation in chemicals and products, information about existing uses can quickly become outdated. More recent and relevant information on chemical handling and product use, if available, can significantly enhance the accuracy of both task exposure and product safety assessments.

Risk assessors often use existing data in the literature, direct monitoring and observation of workers or consumers, or surveys and questionnaires to quantify exposure information from chemical handling or product use and evaluate potential risks. Exposure information includes, for example, the duration of a task or product use, the distance of a chemical from the breathing zone, and the amount of unprotected, exposed skin. While image or video recordings are sometimes utilized in risk assessment, they primarily serve as supplementary tools. Translating these recordings into quantitative or semi-quantitative data requires human review, which is time-consuming, subjective, and not easily scalable.

The advancement and widespread availability of recording devices, such as smartphones and tablets, have made it easy to generate large volumes of pictures and videos. These recordings, which include various work tasks in occupational settings and product use scenarios for consumers, contain valuable exposure information.

Deep learning, a subfield of artificial intelligence and machine learning, is designed to tackle unstructured data like images and videos. The technology has been widely used in many daily applications such as facial recognition, self-driving cars, and medical diagnosis.

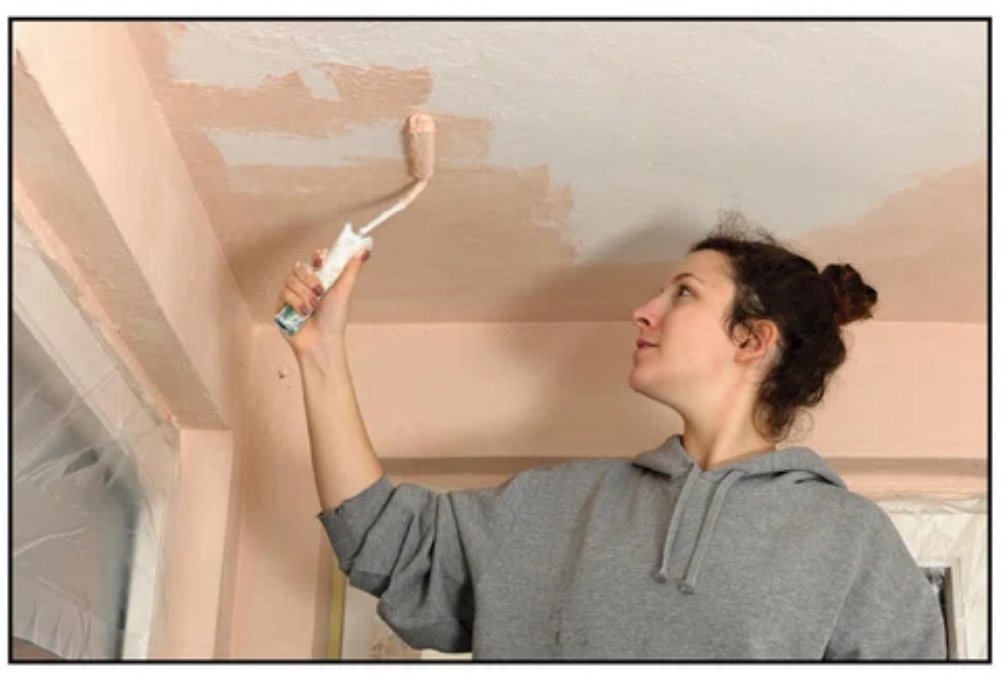

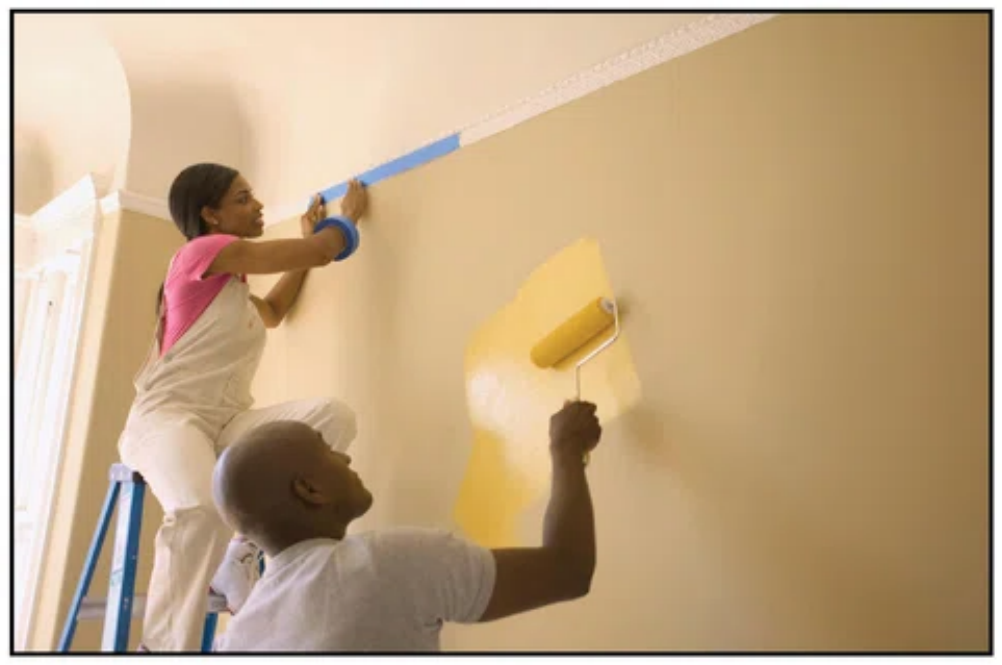

Figure 1. The performance of color-based approaches to computer vision may be affected by the lighting conditions under which photos were taken. In addition, the default skin color spectrum setting may not be broad enough to cover dark skin tones.

To evaluate the capabilities and readiness of deep learning technology to improve dermal exposure estimates, the authors conducted a proof-of-concept study focused on quantification of exposed skin areas from images of indoor painting. Quantifying the surface area of skin that comes into contact with the chemical being assessed is an important aspect of dermal exposure assessments. Often, this parameter is qualitatively estimated based on assumptions such as typical clothing worn or body parts exposed and estimated average surface area of the body parts. For example, the AIHA IH SkinPerm tool has a default exposed skin surface area of 1,000 $cm^2$, which is the estimated surface area of two adult hands. While our study focused on consumer product use, the approach we developed can be applied to workplace tasks where dermal exposure may be a concern. We also evaluated the performance of the computer-based approach by comparing the deep learning results with human estimations.

## TECHNOLOGY DEVELOPMENT AND RESULTS

Computer vision is a field of artificial intelligence that enables computers to detect and interpret images and videos. It has rapidly advanced from traditional color-based approaches to many recent developments that leverage deep learning. When, as is often the case, no off-the-shelf computer vision method exists for specific applications, the solution typically requires transferring learning from existing architectures and then customizing them with relevant images. To train and evaluate our model for this study, we purchased one hundred seventy pictures of indoor painting scenarios from Getty Images, processed them, and analyzed them using different computer vision approaches. To benchmark our model's performance, humans visually identified the exposed body parts in the images and converted them into percentages based on the standard body-part-to-whole-body ratio for adult populations. We used the VGG Image Annotator tool, or VIA, to label pictures in Common Objects in Context (COCO), a standard format for storing and sharing annotations of images and videos. We started with a traditional computer vision method. Unlike deep learning, this method does not require a large number of training images and can detect the color segmentation of objects in pictures based on their color space values. Color spaces are a fundamental concept in image processing and play a crucial role in computer vision. IBM defines a color space as "a representation of the individual colors that can be combined to create other colors." Because devices such as computer and TV monitors, printers, and scanners create colors differently, different methods of describing colors are needed for each device. The color space defines a standardized method of representing colors in digital images, allowing for efficient processing and analysis. RGB (Red Green Blue) is the most commonly used color space in digital imaging, but other color spaces, such as CMYK, HSV, and YUV, are also widely used.

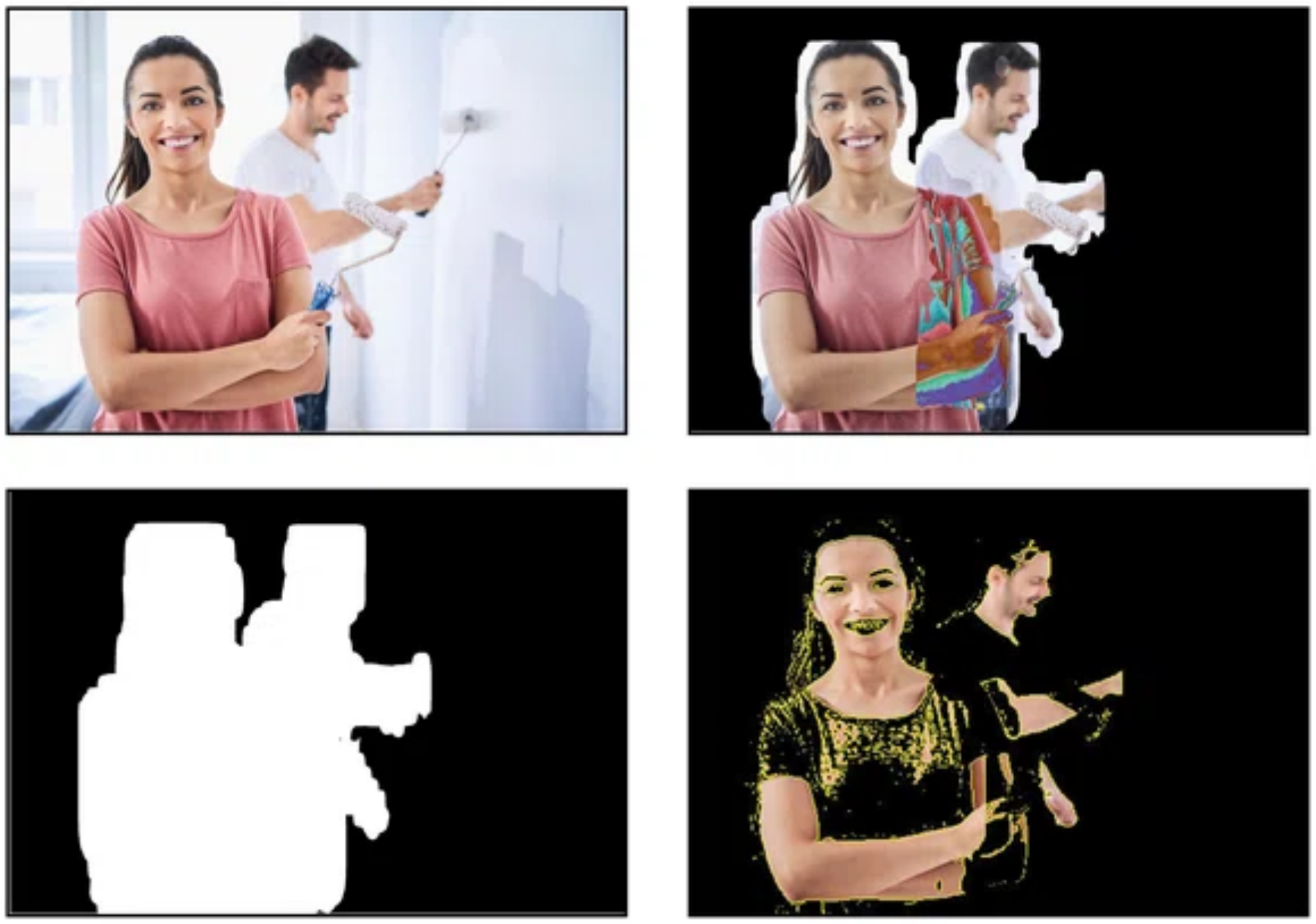

Figure 2. A hybrid computer vision approach identifies human subjects and then detects exposed skin areas in images.

However, the color-based approach has several challenges. For example, its performance is affected by varying lighting conditions when the pictures are taken or if used under its default skin color spectrum setting, which may not be broad enough to cover dark skin tones (see Figure 1). To address these issues, we refined the existing algorithm but determined that it did not work well when the background color was similar to the skin color of the human subject in the image.

We then explored Mask Region-based Convolutional Neural Network (Mask R-CNN), the latest deep learning technology in computer vision. It can identify multiple objects from pictures instantaneously by drawing bounding boxes and contours. Once customized with training images, Mask R-CNN can identify human subjects and their exposed skin areas all at once. However, developing a specific method for skin segmentation would have required us to retrain the existing architecture with as many as ten thousand training images. With a limited number of images available for our pilot study, the performance of the retrained model was limited to identifying exposed skin. Also, the effort of labeling a large number of training images is quite high.

To overcome the limitations and leverage the strengths of the color-based and deep learning methods, we developed a hybrid approach. First, we applied the Mask R-CNN to identify the human subjects from images and then to mask the background based on the outlines of the human subjects identified. This effectively minimized background color interference. Then we applied the color-based approach on the masked images to identify the exposed skin areas from the human subjects in the pictures.

We ran the hybrid approach for all 170 images and evaluated its performance. The trained Mask R-CNN successfully identified human subjects and detected exposed skin areas from all the masked pictures (see Figure 2). We calculated the ratios, in pixels, of the exposed skin areas to human subjects from the hybrid approach and then benchmarked them against estimates made by humans looking at the same images. The overall agreement between the hybrid approach and study volunteers was about 80 percent.

## FURTHER DEVELOPMENT

This proof-of-concept pilot study demonstrated the potential of using AI to extract exposure information from unconventional sources such as images to enhance occupational and consumer exposure assessments. While the study focused on consumer indoor painting, the approach can be applied to different workplace exposures where images and videos provide critical contextual information about how tasks are performed, and the exposure variability is driven by human behaviors.

The hybrid approach offers several advantages. It is not limited by the number of available images and can be easily scaled up for analyzing a large set of pictures. For example, the hybrid approach could be used to provide information about exposed skin from a single photo of a given task in a systematic way; however, the approach can also be used to analyze a larger set of images to generate averages or ranges of typical surface areas of exposed skin. Moreover, this image analysis workflow can be generalized to extract data from pictures to support exposure assessments for other scenarios and tasks.

There are a few areas for further improvement, especially in skin quantification. Currently, the approach only quantifies the percentage of exposed skin relative to the human subjects in the images. Enhancing the accuracy of this method could involve incorporating a feature to identify specific body parts or clothing and personal protective equipment. Additionally, enhancing this workflow to process videos can provide a more comprehensive view of exposure scenarios, capturing time-series exposure profiles and offering a complete picture of the exposure during chemical handling tasks or product uses.

Nevertheless, this project showcased a novel approach of using AI to generate exposure data from unstructured data sources. This method can assist exposure scientists and other OEHS professionals by leveraging key contextual information found in images and videos to improve existing exposure scenarios or evaluate new exposure scenarios for chemical handling and product safety assessments in a relevant and cost-effective manner.

HUA QIAN, PhD, is an exposure scientist and AI lead in health and environmental sciences at ExxonMobil Biomedical Sciences Inc.

MANISHA KOTHA, MS, is a data scientist at ExxonMobil Technology and Engineering Company.

TUAN TRAN, PhD, is an application architect at ExxonMobil Technical Computing Company.

JENNIFER SHIN, MHS, CIH, is an exposure scientist at ExxonMobil Biomedical Sciences Inc. and chair of the AIHA Emerging Digital Technologies Committee.

HAINING ZHENG, PhD, is principal data scientist at ExxonMobil Technology and Engineering Company.